\documentclass[10pt,twocolumn,letterpaper]{article}
\newcommand{\ARXIV}[2]{#2} 

\usepackage[cvpr]{templates/EarthVision/cvpr} 

\usepackage{multirow}
\usepackage{booktabs} 
\usepackage{makecell} 
\usepackage{array}    
\usepackage{geometry}
\usepackage{amsmath}
\usepackage{color}
\usepackage{soul}
\usepackage{svg}
\usepackage{colortbl}
\usepackage{pdflscape}
\usepackage{algorithm}
\usepackage{algorithmic}
\usepackage{adjustbox}
\usepackage{graphicx}
\usepackage{tabularx}
\usepackage{fontawesome5} 
\usepackage{times}
\usepackage{subcaption}
\usepackage{textcomp}
\usepackage{placeins}
\usepackage{tikz}
\usepackage{xurl} 
\usepackage{pifont}

\usetikzlibrary{shapes,patterns}
\usetikzlibrary{arrows.meta}
\usetikzlibrary{fadings, shadows}
\usetikzlibrary{arrows, positioning, fit, backgrounds, calc}
\usetikzlibrary{matrix, fit}

\definecolor{rowhighlight}{HTML}{c6def7} 
\newcommand{\red}[1]{{\color{red}#1}}

\definecolor{cvprblue}{rgb}{0.21,0.49,0.74}

\usepackage{styles}

\def\paperID{9} 
\def\confName{EarthVision}
\def\confYear{2025}

\PassOptionsToPackage{hyphens}{url}
\usepackage{amsmath}
\usepackage{times}
\usepackage{mathtools}
\usepackage{amssymb}
\usepackage{balance}
\usepackage{enumitem}
\usepackage{pgfplots}
\usepackage{microtype}
\usepackage{pgf}
\usepackage{paralist}
\usepackage{ifthen}
\usepackage[hang,flushmargin]{footmisc}
\usepackage[accsupp]{axessibility}  
\usepackage[pagebackref,breaklinks,colorlinks,allcolors=cvprblue]{hyperref}

\pgfplotsset{compat=1.18}

\renewcommand{\paragraph}[1]{\par\vskip 1mm\noindent\textbf{#1}}

\begin{document}

\title{
CoDEx: Combining Domain Expertise for Spatial Generalization \\ in Satellite Image Analysis}

\author{
    Abhishek Kuriyal\textsuperscript{1}
    \and
    Elliot Vincent\textsuperscript{1,2,3}
    \and
    Mathieu Aubry\textsuperscript{1}
    \and
    Loïc Landrieu\textsuperscript{1,2}
    \and
    {\textsuperscript{1}LIGM, ENPC, IP Paris, Univ Gustave Eiffel, CNRS, France}\\
    {\textsuperscript{2}LASTIG, Univ Gustave Eiffel, IGN-ENSG, 94160, Saint-Mande, France}\\
    {\textsuperscript{3}Inria, ENS, CNRS, PSL Research University, France}
}

\maketitle

\begin{abstract}
Global variations in terrain appearance raise a major challenge for satellite image analysis, leading to poor model performance when training on locations that differ from those encountered at test time. This remains true even with recent large global datasets. To address this challenge, we propose a novel domain-generalization framework for satellite images. Instead of trying to learn a single generalizable model, we train one expert model per training domain, while learning experts' similarity and encouraging similar experts to be consistent. A model selection module then identifies the most suitable experts for a given test sample and aggregates their predictions. Experiments on four datasets (DynamicEarthNet, MUDS, OSCD, and FMoW) demonstrate consistent gains over existing domain generalization and adaptation methods. Our code is publicly available at \GITHUB.
\end{abstract}


\section{Introduction}
Earth observation data often exhibits significant spatial domain shifts, such as diverse biomes, architectures, or climate~\cite{vincent2024satellite, scheibenreif2024parameter, tasar2020standardgan}. Consequently, trained models struggle to generalize to new conditions, especially when they differ substantially from the ones of the training data. This shortfall is particularly problematic because models then perform best where annotations are already abundant, rather than in regions where their prediction is most needed. 

\begin{figure}
    \centering
    \input{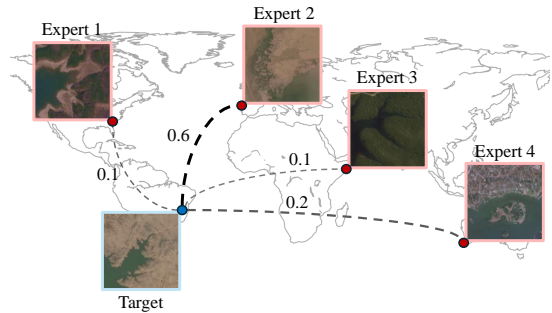}
    \vspace{-1.5em}
    \caption{{\bf Combing Domain Experts.} The appearance of satellite images can vary significantly across regions, even when they contain the same semantic features (here, coastlines). We train experts for each discrete location in the train set, and learn to aggregate the most relevant ones when confronted with an unknown domain.}\vspace{-1em}
    \label{fig:teaser}
\end{figure}

Deep learning models are especially vulnerable to these domain shifts because of their capacity to learn and overfit complex data distributions. Most domain generalization strategies learn a single domain-invariant model~\cite{li2022domain, arjovsky2019invariant, krueger2021out}, but real-world benchmarks (\eg, WILDS~\cite{koh2021wilds}) show that such approaches often fail to surpass simple baselines. Domain adaptation techniques---which rely on unlabeled data from a target domain---face similar hurdles, as shown in the U-WILDS benchmark~\cite{sagawa2022extending}. These findings underscore the difficulty of addressing domain shift for Earth observation, which remains a mostly unsolved problem despite its considerable importance.

In this paper, we introduce CoDEx (COmbining Domain EXperts), a multi-expert approach to domain generalization tailored for satellite images. Instead of training a single model to perform well on all domains, we learn a specialized expert for each domain in the training set, but also a similarity between experts, which we use to encourage the consistency of similar experts to improve robustness. To aggregate experts' predictions, we then train a model selection module to weight the outputs of the domain experts for a given input image, see \cref{fig:teaser}. At inference, when faced with a completely new domain, we use this selection module---without any additional fine-tuning on test data---to combine the predictions of all experts.

We validate our approach on multiple satellite image and time series datasets, including DynamicEarthNet~\cite{toker2022dynamicearthnet}, MUDS~\cite{van2021multi}, OSCD-3ch.~\cite{daudt2018urban} and FMoW~\cite{christie2018functional}, and show consistent performance gains. In summary:
\begin{compactitem}
\item We propose a novel multi-domain training strategy that enforces consistency across domain experts without relying on hand-crafted similarity metrics between domains.
\item  We design a model-selection module to accurately predict which domain experts will yield the most reliable predictions, allowing us to perform robust spatial generalization with minimal computational overhead.
\item  We show consistent qualitative and quantitative improvements over existing domain-generalization and even domain-adaption methods, on four datasets and three baselines.
\end{compactitem}

\section{Related Work}
In this section, we discuss existing work on unsupervised domain adaptation and domain generalization for Earth observation, and multi-experts models.

\paragraph{Domain Adaptation for Earth Observation.} 
The goal of domain adaptation is to modify a model trained on annotated domains to optimize it for a target domain where only unannotated data is available. The availability of unlabeled satellite images makes such an approach particularly attractive to Earth observation~\cite{koh2021wilds, sagawa2022extending, lu2024global, zeng2024domain}. Among the most commonly used methods, feature alignment rely on moment matching \cite{sun2016deep}, discriminative losses \cite{long2018conditional}, or entropy minimization~\cite{vu2019advent, chen2019domain} to adapt a deep learning model to a new domain.
Another approach is to rely on confident predictions in the new domain, for example using pseudo-labeling~\cite{lee2013pseudo}, FixMatch~\cite{sohn2020fixmatch}, or Noisy Student~\cite{xie2020self}. In the case of spatial domain adaptation, spatial awareness can be incorporated into the model by leveraging geographical metadata, as demonstrated in~\cite{marsocci2023geomultitasknet}. StandardGAN~\cite{tasar2020standardgan} and StyleAugment~\cite{chun2021styleaugment} take a different approach by modifying the data distribution itself with image translation. Recently, self-supervised learning has emerged as a powerful tool for domain adaptation, where models are trained on large-scale unlabeled data in a self-supervised way and on labeled data~\cite{manas2021seasonal, saito2020universal, caron2020unsupervised}. In that sense, recent Earth observation foundation models like DOFA~\cite{xiong2024neural}, pretrained on datasets covering a significant portion of the globe, can be seen as domain adaptation tools. 

\paragraph{Domain Generalization for Earth Observation.} Unlike domain adaptation, which relies on unlabeled data from the target domain, generalization approaches must learn robust representations from source data alone. 
The features can be trained to be invariant to spectral and temporal transformations through augmentation like color jittering, image mixing~\cite{olsson2021classmix, illarionova2021mixchannel, he2022classhyper} or manipulations of time series~\cite{nyborg2022timematch, garnot2022multi,yuan2025empirical}. This can also be achieved through the design of the features themselves, such as transformation-invariant prototypes~\cite{vincent2023pixel}, or through loss functions.
For example, contrastive losses align features of spatially or temporally distant images~\cite{mall2023change}, or between an image and its distorted version~\cite{zhao2021contrastive}.
In contrast to these methods that focus on learning a single robust model, our approach trains and combines multiple domain-specific networks, following the idea of multi-expert models.

\begin{figure*}[t]
    \centering
    \begin{tabular}{@{\hspace{-5mm}}cc@{\hspace{1cm}}c} 
        \begin{minipage}{0.25\textwidth}  
            \centering
            \begin{tikzpicture}[scale=0.6, transform shape]
            \input{figures/baseline.tex}
            \end{tikzpicture}
            \subcaption{Baseline}
            \label{fig:over:baseline}
        \end{minipage}
    &
        \begin{minipage}{0.25\textwidth}  
            \centering
            \begin{tikzpicture}[scale=0.6, transform shape]
                \input{figures/multihead.tex}
            \end{tikzpicture}
            \subcaption{Multi-Model}
            \label{fig:over:multi}
        \end{minipage}
    &
        \begin{minipage}{0.4\textwidth} 
            \centering
            \begin{tikzpicture}[scale=0.6, transform shape]
                \input{figures/multiheadwlosses.tex}
            \end{tikzpicture}
            \subcaption{Multi-Model Training}
            \label{fig:over:multiours}
        \end{minipage}
    \end{tabular}
\caption[Overview of methodology]{
{\bf Multi-Domain Training.} 
We present the different multi-domain training approaches explored in this paper. In \subref{fig:over:baseline}, we train a single model on all training domains. In \subref{fig:over:multi}, we train one model per training domain; all models share the same backbone network, and only see data from their domain. In \subref{fig:over:multiours}, we add a consistency loss ensuring that the prediction of models associated to similar domains---as defined by a learnable affinity matrix---also produce accurate results. \multx~indicates vector multiplication and \resizebox{2mm}{!}{\fire} a module with tunable parameters.}
\label{fig:over0}
\end{figure*}

\paragraph{Multi-Experts Models.} Multi-expert models aggregate predictions from multiple expert learners. 
Each expert may specialize in certain domains~\cite{chen2025lfme} or classes~\cite{zermatten2024land}. Experts may also be  encouraged to diversify their predictions through balancing losses and different initializations~\cite{shazeer2017outrageously}. Strategies for combining expert predictions at inference when given an out-of-domain input vary. In particular, D\textsuperscript{3}G~\cite{yao2024improving} aggregates predictions from domain-specific models using weights predicted as a function of domain metadata. Instead, our proposed expert selection module only relies on input features and does not require any metadata at test time.
\section{Method}
We propose CoDEx, a spatial generalization framework for satellite image analysis. Like domain generalization models, we handle domain shifts without retraining on target domain data or requiring domain-specific adapters. Our training set consists of satellite images or time series taken from $D$ distinct source domains, denoted as $\cD_1, \dots, \cD_D$, on which we perform segmentation or classification. We will leverage this multi-domain source data to train a model that generalizes to data from an unseen target domain from a different location.

To achieve this, we first train a set of domain experts (\cref{sec:multihead}) and then introduce a mechanism to select and combine the most relevant model predictions for a given test input (\cref{sec:headselection}). Implementation details are provided in \cref{sec:details}.

\subsection{Multi-Domain Training}
\label{sec:multihead}
We train a set of $D$ models $\phiseg{1}, \dots, \phiseg{D}$, each specialized for a specific training domain $\cD_1, \cdots, \cD_D$, \ie domain experts. These models share the same feature extractor backbone
and each maintains a small, domain-specific segmentation or classification head $h_1, \cdots h_D$.
In a first training stage, visualized in~\cref{fig:over0}, we train the model to produce accurate and domain-consistent predictions with the two loss functions described below.
\paragraph{Domain Loss.}
Let $x$ be an input sample (image or time series) from domain $\cD_d$, with corresponding ground truth label $y$ (class or label map). We supervise the predictions of the expert $\phiseg{d}$ corresponding to domain $\cD_d$ with the following loss applied to inputs $x$ in domain $\cD_d$:
\begin{align}
    \cLseg(x) = \ell(\phiseg{d}(x), y)~,
\label{eq:seg}
\end{align}
where $\ell$ is a standard classification loss. We use the focal loss \cite{ross2017focal} in our experiments, and apply it pixel-wise for segmentation tasks.

\paragraph{Consistency Loss.}
Yao \etal \cite{yao2024improving} propose improving cross-domain generalization by enforcing consistency among the predictions of heads from ``similar'' domains. To quantify domain similarity, they construct an affinity matrix $a \in \mathbb{R}^{D \times D}$, where each entry $a_{d,e}$ encodes the proximity between domains $\cD_d$ and $\cD_e$. For geospatial applications, they compute $a$ as a combination of handcrafted and learned components. Specifically, they set the handcrafted part of $a_{d,e}$ to $1$ if domains $\cD_d$ and $\cD_e$ are in adjacent regions and $0$ otherwise. The learned part is a learned function of  geographical metadata.

Instead, we propose to learn the affinity matrix $a$ directly. More precisely, we define $a\in \mathbb{R}^{D \times D}$ as the row-wise softmax---excluding the diagonal coefficients---of a learnable parameter matrix in $\mathbb{R}^{D \times D}$. We encourage consistency between domains by minimizing the following loss for an input $x$ in a domain $\cD_d$:
\begin{align}
    \cLcon(x) = \ell( \sum_{e \neq d} a_{d,e} \phiseg{e}(x), y )~.
\label{eq:con}
\end{align}
This loss encourages predictions from domains $\cD_e$ close to $\cD_d$ according to $a$, to also be effective for $x$ sampled from $\cD_d$. Unlike $\cLseg$, which updates only the parameters of the expert expert $\phiseg{d}$ for inputs in $\cD_d$, the consistency loss $\cLcon$ propagates gradients to all $\phiseg{e}$ for $e \neq d$. Compared to Yao \etal \cite{yao2024improving}, this not only provides greater flexibility but also eliminates the manual design process and hyperparameters related to the definition of the hand-crafted part of $a$. Our ablation study also shows that it yields better results.

\begin{figure}[t]
    \centering
        \begin{minipage}{1\linewidth}  
            \centering
            \begin{tikzpicture}[scale=0.6, transform shape]
                \input{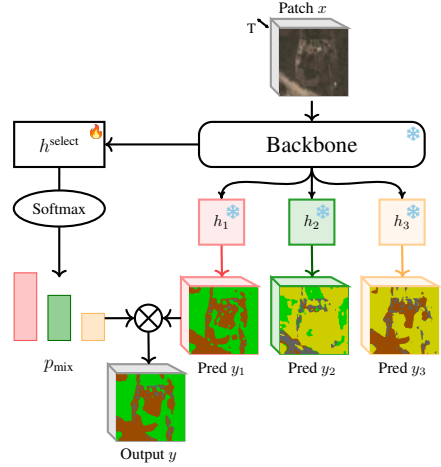}
            \end{tikzpicture}
        \end{minipage}
        \vspace{-2mm}
\caption{{\bf Domain Expert Selection}. We freeze the models trained previously, and train a domain expert selection model to select the most relevant models for a given input sample from an unseen domain.  \raisebox{0.5mm}{\resizebox{2.5mm}{!}{\multx}}:~ vector multiplication, \textcolor{lightblue}{\resizebox{2mm}{!}{\faSnowflake}}: frozen layers,  \resizebox{2mm}{!}{\fire}: module with tunable parameters.}
\label{fig:over}
\end{figure}

\subsection{Domain Expert Selection}
\label{sec:headselection}

In a second stage of training, visualized in~\cref{fig:over}, we freeze all models $\phiseg{d}$, and only train an expert selection head $\hsel$ to predict weights to aggregate the predictions of the domain experts. For a given input $x$, $\hsel$ leverages features from $x$ provided by the backbone and outputs a vector $\phisel(x)\in\mathbb{R}^D$. We train $\phisel$ with two losses: $\cLacc$, which encourages $\phisel$ to predict the accuracy of experts, and $\cLmix$, which aims at maximizing the quality of a mixture. Note that this module is also exclusively trained with data from the source domains. 

\paragraph{Accuracy Prediction Loss.} We encourage the expert selection $\phisel$ to predict the accuracy of the predictions of each domain-specific expert $\phiseg{d}$ on each sample $x$:
\begin{align}
   \!\!\cLacc(x)\!=\! \sum_d \left\vert \phisel(x)[d] - \acc\left(\phiseg{d}(x), y\right) \right\vert~,
\label{eq:acc}
\end{align}
where $\acc(z,y)$ is a measure of the performance of the prediction $z$ against the ground truth $y$, and $\phisel(x)[d]$ is the component $d$ of $\phisel(x)$. We use pixel-wise overall accuracy for semantic segmentation tasks, and a binary value $0/1$ for classification.

\begin{table*}[t]
    \caption{{\bf Datasets Characteristics}. We choose four datasets of satellite images or time series, and report their resolution, the nature of their annotations, and the number of samples and domains for the training, validation, and test sets. *We create our own custom splits for these datasets to ensure the separation of spatial domains.}
   \centering
    \resizebox{.90\linewidth}{!}{
    \small
    \begin{tabular}{lcccc}
        \toprule
        & \textbf{DynEarthNet}& \textbf{MUDS*} & \textbf{OSCD-3 ch.} & \textbf{FMoW*} \\
        \midrule
        Task & Segmentation & Segmentation & Change detection & Classification\\ 
        Temporal Resolution & 24 (monthly) & 24 (monthly) & 2 (image pair) & 1 (monodate) \\
        Spatial Resolution & 128$\times$128, 3-4m/pix & 128$\times$128, 3-4m/pix& 75$\times$75, 10m/pix& 224$\times$224, 0.3-1.6m/pix \\
        Spectral Resolution & 4: RGB + NIR & 3: RGB & 3: RGB & 3: RGB \\
        Classes & 7 (land cover) & 2 (land cover) & 2 (change) & 62 (land use) \\
        Training & 3520 (55 dom.) & 2560 (40 dom.)  &  896  (14 dom.)  & 59,443 (100 dom.)  \\
        Validation & \hphantom{3}640 (10 dom.)  & \hphantom{3}640 (10 dom.) &  -  & 26,697 (50 dom.)\hphantom{1} \\
        Test & \hphantom{3}640 (10 dom.) & \hphantom{3}640 (10 dom.) &  \hphantom{3}640 (10 dom.) & 32,745 (50 dom.)\hphantom{1} \\
        \bottomrule
    \end{tabular}
    }

    \label{tab:dataset_comparison}
\end{table*}

\paragraph{Mixture Supervision Loss.} 
We convert the predicted accuracy $\phisel(x)$ into a weight vector $\pmix(x) \in \mathbb{R}^{D}_+$ using a softmax with temperature $\tau > 0$, and use it to compute a mixture of predictions from all experts $\sum_d \pmix(x)[d] \phiseg{d}$, which we use at inference. We then define the loss $\cLmix$ as:
\begin{align}
    &\cLmix(x) = \ell\left( \sum_d \pmix(x)[d] \phiseg{d}(x), y \right)~ ,
\end{align}
with $\pmix(x) = \text{softmax}\left( \frac1\tau \phisel(x) \right)$ 
 and $\ell$ the same loss as in~\cref{eq:con,eq:seg}. Note that at inference, we can use the mixture prediction $\sum_d \pmix(x)[d] \phiseg{d}$ without knowing the geographical location of the test sample $x$.

\subsection{Implementation Details}
\label{sec:details}

Our default backbone for satellite image time series (SITS) segmentation and change detection, is a MultiUTAE~\cite{vincent2024satellite,garnot2021panoptic}, which maps SITS to time series of feature maps of the same spatial and temporal resolution as the input and has 260K parameters. The domain heads $h_d$ are $3\times3$ convolutions applied to the highest-resolution feature map of the UTAE for each date for segmentation. The selection head $\hsel$ is also a $3\times3$ convolution that takes as input features from different encoder blocks concatenated along the channel dimension which are then pooled spatially and temporally. To perform change detection, we subtract the logits predicted for the two consecutive images and supervise it with the true binary change map.  

For image classification, we use a larger version of MultiUTAE with 3.1M parameters as the default backbone. Both domain heads $h_d$ and selection head $\hsel$ are linear layers. The domain head is applied to the pooled features, the selection head to features pooled from multiple blocks. 
The affinity matrix adds $D^2$ parameters, which is significantly smaller than the number of parameters in the encoder and heads.


As our results appear to be robust to loss weighting, we always use a weight of $1$ when adding $\cLseg$ and $\cLcon$ in the first stage of training, and $\cLacc$ and $\cLmix$ in the second. The temperature $\tau$ of the softmax in the mixture definition is also set to $1$.

We use the Adam optimizer with a learning rate of \(1 \times 10^{-4}\) and coefficients \((\beta_{1}, \beta_{2}) = (0.9, 0.999)\). Additionally, we apply a weight decay of 0.01. 

\section{Experiments}
In \cref{sec:data}, we present the four Earth observation datasets on which we evaluate our proposed approach for three tasks (classification, semantic segmentation, and change detection), with either single image or time series inputs. We then compare our approach to a wide array of domain-generalization and adaptation methods in \cref{sec:results}. Finally, we ablate and analyze our approach in \cref{sec:ablation}.

\subsection{Datasets and Evaluation}
\label{sec:data}
\begin{table*}
   \caption{{\bf Quantitative Results.} We report the results of various domain generalization and adaptation methods on four datasets. Cell color represents the difference with the baseline model with the following colormap:  
    \protect\tikz[baseline=-0.0em]{\protect\shade[left color=red, right color=white] 
    (0,0) rectangle (1,0.25);\protect\shade[left color=white, right color=green] 
    (1,0) rectangle (2,0.25);\protect\node at (0.20,0.125) {\footnotesize \bf -3};\protect\node at (1,0.125) {\footnotesize \bf +0};\protect\node at (1.80,0.125) {\footnotesize \bf +3};}. \textcolor{gray!80!black}{Oracle} selects the best training head for each test input. $^*$D\textsuperscript{3}G~\cite{yao2024improving} is our own implementation, which we will release. }
     \label{tab:main_results}
    \renewcommand{\arraystretch}{1.05}
     \setlength{\fboxsep}{1.5pt}

    \centering
    \resizebox{\linewidth}{!}{
    \begin{tabular}{@{}l >{\centering\arraybackslash}p{1.4cm} >{\centering\arraybackslash}p{1.4cm} >{\centering\arraybackslash}p{1.4cm} >{\centering\arraybackslash}p{1.4cm} >{\centering\arraybackslash}p{2.8cm} >{\centering\arraybackslash}p{2.8cm}}
        \toprule
        \multirow{2}{*}{\textbf{Method}} & \multicolumn{2}{c}{\textbf{DynamicEarthNet}} & \multicolumn{2}{c}{\textbf{MUDS}}  &  \multicolumn{1}{c}{\textbf{OSCD-3 ch.}} & \multicolumn{1}{c}{\textbf{FMoW}}\\
        \cmidrule(lr){2-3} \cmidrule(lr){4-5} \cmidrule(lr){6-6} \cmidrule(lr){7-7}
         & {mIoU} & {OA} & {mIoU} & {OA} & {F1} & {Avg. Acc.} \\ 
        \midrule
        Baseline \cite{vincent2024satellite}  & \applycolorA{37.5} & \applycolorB{73.8} & \applycolorC{63.5} & \applycolorD{95.1} & \applycolorE{46.5} & \applycolorG{51.6} \\ 
        \midrule
        \rowcolor{gray!80!black!10} \multicolumn{7}{l}{\textbf{Domain Adaptation:} fine-tuned on the test set's input data} \\ 
        StyleAugment \cite{chun2021styleaugment} & \applycolorA{36.6} & \applycolorB{73.4} & \applycolorC{63.1} & \applycolorD{94.7} & \applycolorE{46.3} & \applycolorG{51.3} \\ 
        CORAL \cite{sun2016deep}  & \applycolorA{38.3} & \applycolorB{74.6} & \applycolorC{63.3} & \applycolorD{95.1} & \applycolorE{46.6} & \applycolorG{52.0} \\ 
        Pseudo labels \cite{lee2013pseudo} & \applycolorA{37.4} & \applycolorB{74.1} & \applycolorC{63.7} & \applycolorD{95.0} & \applycolorE{46.7} & \applycolorG{51.7} \\ 
        AdvEnt \cite{vu2019advent} & \applycolorA{36.3} & \applycolorB{73.5} & \applycolorC{62.9} & \applycolorD{94.9} & \applycolorE{45.4} & \applycolorG{51.9} \\ 
        DANN \cite{long2018conditional} & \applycolorA{35.8} & \applycolorB{73.9} & \applycolorC{63.3} & \applycolorD{95.1} & \applycolorE{45.9} & \applycolorG{51.8} \\ 
        MaxSquare \cite{chen2019domain} & \applycolorA{38.1} & \applycolorB{74.5} & \applycolorC{64.0} & \applycolorD{95.2} & \applycolorE{47.1} & \applycolorG{52.1} \\ 
        \midrule
        \rowcolor{gray!80!black!10} \multicolumn{7}{l}{\textbf{Foundation Model-Based:} pretrained on external data, fine-tuned on the train set} \\ 
        DOFA \cite{xiong2024neural} & \applycolorA{35.8} & \applycolorB{73.7} & \applycolorC{63.2} & \applycolorD{94.3} & \applycolorE{46.2} & \applycolorG{51.3} \\ 
        \midrule
        \rowcolor{gray!80!black!10} \multicolumn{7}{l}{\textbf{Domain Generalization:} only sees the train set} \\ 
        ClassMix \cite{olsson2021classmix} & \applycolorA{36.2} & \applycolorB{74.1} & \applycolorC{63.2} & \applycolorD{94.8} & \applycolorE{46.8} & \applycolorG{51.5} \\ 
        Contrastive Seg. \cite{zhao2021contrastive} & \applycolorA{37.9} & \applycolorB{73.6} & \applycolorC{63.3} & \applycolorD{95.0} & \applycolorE{46.4} & \applycolorG{53.0} \\
        D\textsuperscript{3}G~\cite{yao2024improving}$^*$ & \applycolorA{38.7} & \applycolorB{75.2} & \applycolorC{63.9} & \applycolorD{95.2} & \bf\applycolorE{48.1} & \applycolorG{53.5} \\ 
        \bf CoDEx (Ours) & \bf\applycolorA{39.1} & \textbf{\applycolorB{75.7}} & \bf\applycolorC{64.2} & \bf\applycolorD{95.8} & \applycolorE{47.8} & \bf\applycolorG{53.9} \\
        \midrule
        \textcolor{gray!80!black}{Oracle} & \textcolor{gray!80!black}{48.1} & \textcolor{gray!80!black}{81.6}& \textcolor{gray!80!black}{65.8}& \textcolor{gray!80!black}{95.6}& \textcolor{gray!80!black}{56.9}& \textcolor{gray!80!black}{91.1}\\
        \bottomrule
    \end{tabular}
    }

\end{table*}

\paragraph{Datasets.}
We build our benchmark for domain generalization from four standard SITS datasets, which can be naturally split into distinct spatial domains. We provide details about these dataset, DynamicEarthNet~\cite{toker2022dynamicearthnet}, MUDS~\cite{van2021multi}, OSCD-3ch.~\cite{daudt2018urban}, and FMoW~\cite{christie2018functional}, in \cref{tab:dataset_comparison}.
The domains are defined by the unique geographic locations of the images, except for FMoW for which spatial domains correspond to countries. We use train/val/test splits such that the domains are separated. 
We split the large spatial extent of the SITS of DynamicEarthNet, MUDS, and OSCD-3ch. into smaller SITS of spatial size $128\times128$, $128\times128$, and $75\times75$ respectively. For DynamicEarthNet, we only consider the images of the time series that are annotated: one image every month. For FMoW, we treat each image of the time series independently as in WILDS~\cite{koh2021wilds}.

We evaluate model performance with the class-average intersection over union (mIoU) and overall accuracy (OA) for semantic segmentation, F1 score for change detection, and average accuracy (Avg. Acc.) for classification.

\paragraph{Competing Methods.} We evaluate for each dataset a baseline approach, which trains a single model regardless of domains (see \cref{fig:over:baseline}). We evaluated for the first time on these datasets $6$ domain-adaptation methods (StyleAugment~\cite{chun2021styleaugment}, CORAL~\cite{sun2016deep}, Pseudo labels~\cite{lee2013pseudo}, Advent~\cite{vu2019advent}, DANN~\cite{long2018conditional}, MaxSquare~\cite{chen2019domain}) and $3$ domain-generalization approaches (ClassMix~\cite{olsson2021classmix}, Contrastive Seg.~\cite{zhao2021contrastive}, D\textsuperscript{3}G~\cite{yao2024improving}).
Domain-adaptation methods are fine-tuned with the input data from the test set and without labels. We also evaluated a foundation model for Earth observation, DOFA~\cite{xiong2024neural}. As this model takes only a single image as input, we evaluate it for each image independently. Since no implementation of D\textsuperscript{3}G~\cite{yao2024improving} is available, results are from our own implementation, which we will release. For the other methods, we modify the official code to apply it to our data. 
\subsection{Results}\label{sec:results}

\begin{figure*}[t!]
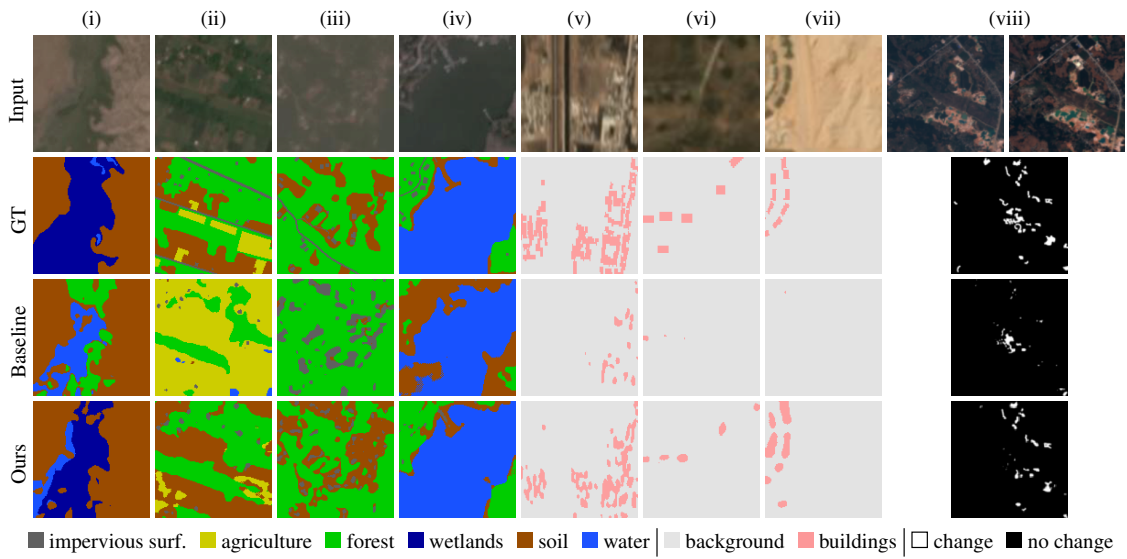

\definecolor{IMPERVCOLOR}{RGB}{96, 96, 96}
\definecolor{AGRI}{RGB}{204, 204, 0}
\definecolor{FOR}{RGB}{0, 204, 0}
\definecolor{WET}{RGB}{0, 0, 153}
\definecolor{SOIL}{RGB}{153, 76, 0}
\definecolor{WATER}{RGB}{30, 83, 255}
\definecolor{BKG}{RGB}{228, 228, 228}
\definecolor{BUILD}{RGB}{252, 152, 153}
\centering
\setlength{\tabcolsep}{1pt}
\resizebox{\linewidth}{!}{
    \begin{tabular}{ccccccccccc}
        & {\scriptsize (i)} & {\scriptsize (ii)} & {\scriptsize (iii)} & {\scriptsize (iv)} & {\scriptsize (v)} & {\scriptsize (vi)} & {\scriptsize (vii)} & \multicolumn{2}{c}{{\scriptsize (viii)}}\\ 
        \rotatebox[origin=c]{90}{\scriptsize Input} & \satimgsetuptwo{input}
        \vspace{2pt}\\
        \rotatebox[origin=c]{90}{\scriptsize GT} & 
        \satimgsetupone{gt}
        \vspace{2pt}\\
        \rotatebox[origin=c]{90}{\scriptsize Baseline} & \satimgsetupone{baseline}
        \vspace{2pt}\\
        \rotatebox[origin=c]{90}{\scriptsize Ours} & \satimgsetupone{ours}
        \vspace{2pt}\\
        \multicolumn{10}{c}{
        \footnotesize{\begin{tabular}{rl@{\;\;}rl@{\;\;} rl@{\;\;} rl@{\;\;} rl@{\;\;} rl @{\;}|@{\;} rl @{\;\;}rl @{\;}|@{\;} rl @{\;\;}rl}
       \tikz{\fill[fill=IMPERVCOLOR, scale=0.18, draw=none] (0,0) rectangle (1,1);} &\scriptsize{impervious surf.}&
       \tikz{\fill[fill=AGRI, scale=0.18, draw=none] (0,0) rectangle (1,1);} &\scriptsize{agriculture}&
       \tikz{\fill[fill=FOR, scale=0.18, draw=none] (0,0) rectangle (1,1);} &\scriptsize{forest}&
       \tikz{\fill[fill=WET, scale=0.18, draw=none] (0,0) rectangle (1,1);} &\scriptsize{wetlands}&
       \tikz{\fill[fill=SOIL, scale=0.18, draw=none] (0,0) rectangle (1,1);} &\scriptsize{soil}&
       \tikz{\fill[fill=WATER, scale=0.18, draw=none] (0,0) rectangle (1,1);} &\scriptsize{water}&
       \tikz{\fill[fill=BKG, scale=0.18, draw=none] (0,0) rectangle (1,1);} &\scriptsize{background}&
       \tikz{\fill[fill=BUILD, scale=0.18, draw=none] (0,0) rectangle (1,1);} &\scriptsize{buildings}&
       \tikz{\fill[fill=white, scale=0.18, draw=black] (0,0) rectangle (1,1);} &\scriptsize{change}&
       \tikz{\fill[fill=black, scale=0.18, draw=none] (0,0) rectangle (1,1);} &\scriptsize{no change}
        \end{tabular}}
        }
    \end{tabular}}
\vspace{-.7em}
    \caption{{\bf Qualitative Segmentations.} We illustrate for random patches the predictions of our method and our baseline. Images from (i-iv) are selected from DynamicEarthNet, (v-vii) from MUDS, and (viii) from OSCD-3ch.}
    \label{fig:qualitative}
\end{figure*}

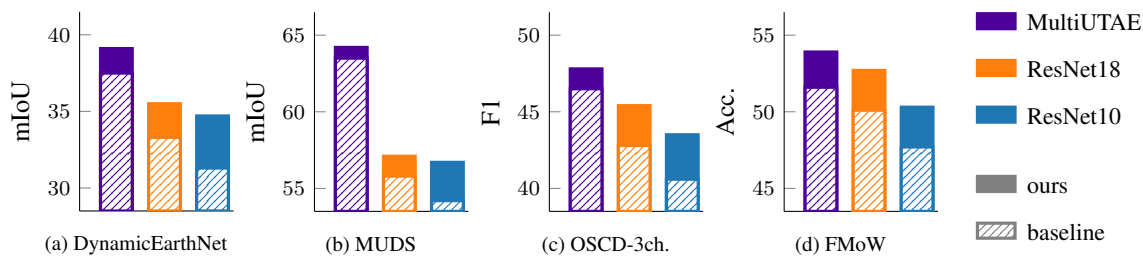
\begin{figure*}[ht]
    \centering
\definecolor{MultiUTAE}{RGB}{76, 0, 153}        
\definecolor{MultiUTAE_LIGHT}{RGB}{152, 102, 255} 
\definecolor{ResNet10}{RGB}{31, 119, 180}     
\definecolor{ResNet10_LIGHT}{RGB}{174, 199, 232} 
\definecolor{ResNet18}{RGB}{255, 127, 14}     
\definecolor{ResNet18_LIGHT}{RGB}{255, 180, 90}  

\begin{tabular}{@{\hspace{-8mm}}c@{\hspace{1mm}}c@{\hspace{1mm}}c@{\hspace{1mm}}c@{\hspace{3mm}}c@{}}
\begin{minipage}[t][2.2cm][t]{.2\linewidth}
\begin{tikzpicture}
    \begin{axis}[
        title style={align=center, font=\footnotesize},
        width=1.2\linewidth,
        height=1.4\linewidth,
        title={},
        ylabel={mIoU},
        ylabel style={font=\footnotesize, yshift=3pt},
        ylabel near ticks,
        ybar,
        bar width=12pt,
        enlarge y limits=0.15,
        enlarge x limits=0.3,
        symbolic x coords={MultiUTAE, ResNet18, ResNet10},
        xtick={MultiUTAE, ResNet18, ResNet10},
        xticklabels={,,},
        x tick label style={rotate=45, anchor=east, font=\footnotesize},
        y tick label style={font=\footnotesize},
        ytick={30, 35, 40},
        ymin=30,
        ymax=40,
        bar shift=3pt,
        axis y line*=left,
        axis x line*=bottom,
        xmajorticks=false
    ] 
    \addplot[fill=MultiUTAE, draw=MultiUTAE, very thick] coordinates {(MultiUTAE,39.1)};
    \addplot[fill=white] coordinates {(MultiUTAE,37.5)}; 
    \addplot[pattern=north east lines, pattern color=MultiUTAE,draw=MultiUTAE, very thick] coordinates {(MultiUTAE,37.5)}; 

    \addplot[fill=ResNet18,draw=ResNet18, very thick] coordinates {(ResNet18,35.5)};
    \addplot[fill=white] coordinates {(ResNet18,33.3)};
    \addplot[pattern=north east lines, pattern color=ResNet18,draw=ResNet18, very thick] coordinates {(ResNet18,33.3)};

    \addplot[fill=ResNet10,draw=ResNet10, very thick] coordinates {(ResNet10,34.7)};
    \addplot[fill=white] coordinates {(ResNet10,31.3)};
    \addplot[pattern=north east lines, pattern color=ResNet10,draw=ResNet10, very thick] coordinates {(ResNet10,31.3)};
    \end{axis}
    \node[font=\footnotesize, align=center] at (0.75,-0.45) {(a) DynamicEarthNet};
\end{tikzpicture}
\end{minipage}
&
\begin{minipage}[t][2.2cm][t]{.2\linewidth}
\begin{tikzpicture}
    \begin{axis}[
        title style={align=center, font=\footnotesize},
        width=1.2\linewidth,
        height=1.4\linewidth,
        title={},
        ylabel={mIoU},
        ylabel style={font=\footnotesize, yshift=3pt},
        ylabel near ticks,
        ybar,
        bar width=12pt,
        enlarge y limits=0.15,
        enlarge x limits=0.3,
        symbolic x coords={MultiUTAE, ResNet18, ResNet10},
        xtick={MultiUTAE, ResNet18, ResNet10},
        xticklabels={,,},
        x tick label style={rotate=45, anchor=east, font=\footnotesize},
        y tick label style={font=\footnotesize},
        ytick={55, 60, 65},
        ymin=55,
        ymax=65,
        bar shift=3pt,
        axis y line*=left,
        axis x line*=bottom,
        xmajorticks=false,
    ] 
    \addplot[fill=MultiUTAE, draw=MultiUTAE, very thick] coordinates {(MultiUTAE,64.2)};
    \addplot[fill=white] coordinates {(MultiUTAE, 63.5)};
    \addplot[pattern=north east lines, pattern color=MultiUTAE,draw=MultiUTAE, very thick] coordinates {(MultiUTAE, 63.5)}; 

    \addplot[fill=ResNet18,draw=ResNet18, very thick] coordinates {(ResNet18,57.1)};
    \addplot[fill=white] coordinates {(ResNet18, 55.8)};
    \addplot[pattern=north east lines, pattern color=ResNet18,draw=ResNet18, very thick] coordinates {(ResNet18, 55.8)};

    \addplot[fill=ResNet10,draw=ResNet10, very thick] coordinates {(ResNet10, 56.7)};
    \addplot[fill=white] coordinates {(ResNet10, 54.2)};
    \addplot[pattern=north east lines, pattern color=ResNet10,draw=ResNet10, very thick] coordinates {(ResNet10, 54.2)};
    \end{axis}
    \node[font=\footnotesize, align=center] at (0.75,-0.45) {(b) MUDS};
\end{tikzpicture}
\end{minipage}
&
\begin{minipage}[t][2.2cm][t]{.2\linewidth}
\begin{tikzpicture}
    \begin{axis}[
        title style={align=center, font=\footnotesize},
        width=1.2\linewidth,
        height=1.4\linewidth,
        title={},
        ylabel={F1},
        ylabel style={font=\footnotesize, yshift=3pt},
        ylabel near ticks,
        ybar,
        bar width=12pt,
        enlarge y limits=0.15,
        enlarge x limits=0.3,
        symbolic x coords={MultiUTAE, ResNet18, ResNet10},
        xtick={MultiUTAE, ResNet18, ResNet10},
        xticklabels={,,},
        x tick label style={rotate=45, anchor=east, font=\footnotesize},
        y tick label style={font=\footnotesize},
        ytick={40, 45, 50},
        ymin=40,
        ymax=50,
        bar shift=3pt,
        axis y line*=left,
        axis x line*=bottom,
        xmajorticks=false,
    ] 
    \addplot[fill=MultiUTAE, draw=MultiUTAE, very thick] coordinates {(MultiUTAE,47.8)};
    \addplot[fill=white] coordinates {(MultiUTAE,46.5)};
    \addplot[pattern=north east lines, pattern color=MultiUTAE,draw=MultiUTAE, very thick] coordinates {(MultiUTAE,46.5)}; 

    \addplot[fill=ResNet18,draw=ResNet18, very thick] coordinates {(ResNet18, 45.4)};
    \addplot[fill=white] coordinates {(ResNet18,42.8)};
    \addplot[pattern=north east lines, pattern color=ResNet18,draw=ResNet18, very thick] coordinates {(ResNet18, 42.8)};

    \addplot[fill=ResNet10,draw=ResNet10, very thick] coordinates {(ResNet10, 43.5)};
    \addplot[fill=white] coordinates {(ResNet10, 40.6)};
    \addplot[pattern=north east lines, pattern color=ResNet10,draw=ResNet10, very thick] coordinates {(ResNet10, 40.6)};
    \end{axis}
    \node[font=\footnotesize, align=center] at (0.75,-0.45) {(c) OSCD-3ch.};
\end{tikzpicture}
\end{minipage}
&
\begin{minipage}[t][2.2cm][t]{.2\linewidth}
\begin{tikzpicture}
    \begin{axis}[
        title style={align=center, font=\footnotesize},
        width=1.2\linewidth,
        height=1.4\linewidth,
        title={},
        ylabel={Acc.},
        ylabel style={font=\footnotesize, yshift=3pt},
        ylabel near ticks,
        ybar,
        bar width=12pt,
        enlarge y limits=0.15,
        enlarge x limits=0.3,
        symbolic x coords={MultiUTAE, ResNet18, ResNet10},
        xtick={MultiUTAE, ResNet18, ResNet10},
        xticklabels={,,},
        x tick label style={rotate=45, anchor=east, font=\footnotesize},
        y tick label style={font=\footnotesize},
        ytick={45, 50, 55},
        ymin=45,
        ymax=55,
        bar shift=3pt,
        axis y line*=left,
        axis x line*=bottom,
        xmajorticks=false,
    ] 
    \addplot[fill=MultiUTAE, draw=MultiUTAE, very thick] coordinates {(MultiUTAE,53.9)};
    \addplot[fill=white] coordinates {(MultiUTAE,51.6)};
    \addplot[pattern=north east lines, pattern color=MultiUTAE,draw=MultiUTAE, very thick] coordinates {(MultiUTAE,51.6)}; 

    \addplot[fill=ResNet18,draw=ResNet18, very thick] coordinates {(ResNet18,52.7)};
    \addplot[fill=white] coordinates {(ResNet18,50.1)};
    \addplot[pattern=north east lines, pattern color=ResNet18,draw=ResNet18, very thick] coordinates {(ResNet18,50.1)};

    \addplot[fill=ResNet10,draw=ResNet10, very thick] coordinates {(ResNet10, 50.3)};
    \addplot[fill=white] coordinates {(ResNet10, 47.7)};
    \addplot[pattern=north east lines, pattern color=ResNet10,draw=ResNet10, very thick] coordinates {(ResNet10, 47.7)};
    \end{axis}
    \node[font=\footnotesize, align=center] at (0.75,-0.45) {(d) FMoW};
\end{tikzpicture}
\end{minipage}
&
\begin{minipage}[t][2.2cm][t]{.1\linewidth}
\centering
\raisebox{1.7cm}{
\setlength{\tabcolsep}{2pt}
\begin{tabular}{rl}
\tikz{\fill[MultiUTAE, draw=MultiUTAE, very thick] (0,0) rectangle (0.5,0.2);} & \small MultiUTAE \\[2mm]
\tikz{\fill[ResNet18, draw=ResNet18, very thick] (0,0) rectangle (0.5,0.2);} & \small ResNet18 \\[2mm]
\tikz{\fill[ResNet10, draw=ResNet10, very thick] (0,0) rectangle (0.5,0.2);} & \small ResNet10 \\[5mm]
\tikz{\fill[gray, draw=gray, very thick] (0,0) rectangle (0.5,0.2);} & \small ours \\[2mm]
\tikz{\fill[pattern=north east lines, pattern color=gray, draw=gray, very thick] (0,0) rectangle (0.5,0.2);} & \small baseline
\end{tabular}
}
\end{minipage}
\end{tabular}
\vspace{-4mm}
    \caption{{\bf Impact of Backbone.} We report the performance of the baseline and our approach for all four datasets and three backbones networks.}
    \label{fig:backbone}
\end{figure*}

\paragraph{Generalization Performance.}
We report the performance of CoDEx and all competing approaches on all four datasets in \cref{tab:main_results}. Our method consistently outperforms competing methods across all metrics, except D\textsuperscript{3}G, which is slightly better on OSCD-3ch. Notably, only our method and D\textsuperscript{3}G exceed the baseline performance for every dataset, and our approach shows slight improvement over D\textsuperscript{3}G on the other 3 datasets. Interestingly, generalization-focused methods often outperform domain adaptation methods, even though the latter have access to data from the target domain.

We also measure the performance of a \textit{Domain Oracle}, which selects the best-performing expert among $\phiseg{1}, \cdots, \phiseg{D}$ for each test sample. This gives an upper bound on the performance one could achieve by selecting a single expert. On MUDS, the oracle’s performance is only marginally higher than ours, hinting that our expert selection mechanism effectively identifies the best experts. On DynamicEarthNet and OSCD-3ch., the oracle substantially outperforms our approach, although its overall performance remains low, suggesting a considerable domain gap between training and testing. On FMoW, the oracle performs extremely well. This is not surprising, as it takes the best prediction among 100 experts for a classification with 62 classes. 

\paragraph{Qualitative Results.}
We illustrate in \cref{fig:qualitative} the predictions from the baseline approach and our proposed method on three datasets: DynamicEarthNet (i–iv), FMoW (v–vii), and OSCD-3ch. (viii). In (i), our approach correctly resolves an ambiguous image of wetland, which might otherwise be confused with water and forest. 
It also provides more precise segmentation of forest and sediment which might be misclassified as agriculture  (ii). For MUDS, we observe higher recall for buildings in both dense (v) and sparse (vi, vii) regions. For OSCD-3ch. (viii), our method generally improves change detection, although the task remains challenging.

\subsection{Analysis}
\label{sec:ablation}
\paragraph{Ablations.}
We evaluate the contribution of different parts of our approach through ablations on DynamicEarthNet and MUDS in \cref{tab:ablation}. First, rather than predicting a single mixture $\pmix$ across all time steps, we remove temporal pooling of the backbone features and allow the method to choose a different model at each time step. This added flexibility reduces performance, likely because the less informative single-image features make model selection less reliable. Second, we remove either $\cLcon$ or $\cLmix$, which leads to a small decrease of performance. Removing both leads to a significant drop, indicating that enforcing consistency among expert models and supervising the mixture coefficient through the final mixed predictions are both crucial and complementary. Lastly, removing $\cLacc$ and only supervising $\phisel$ with $\cLmix$ leads to a small but consistent performance decrease, suggesting that accuracy is a useful proxy for the relevancy of a domain expert. 

\begin{table}[t]
\caption{{\bf Ablation Study.} We evaluate the impact of several of our design choices.}
\label{tab:ablation}
\setlength{\tabcolsep}{4pt}
\small{
\centering
\begin{tabular}{c@{\;}c@{\;}c@{\;}c cc cc}
\toprule
 \textbf{Temp.} & \multirow{2}{*}{\textbf{$\cLcon$}} & \multirow{2}{*}{\textbf{$\cLmix$}} & \multirow{2}{*}{\textbf{$\cLacc$}} & \multicolumn{2}{c}{\textbf{DynEarthNet}} & \multicolumn{2}{c}{\textbf{MUDS}} \\
\cmidrule(lr){5-6} \cmidrule(lr){7-8}
\textbf{pooling}& & & & \textbf{mIoU} & \textbf{OA} & \textbf{mIoU} & \textbf{OA} \\
\midrule
\ding{52} & \ding{52} & \ding{52} & \ding{52} & \textbf{39.1} & \textbf{75.7} & \textbf{64.2} & \textbf{95.8} \\
\ding{52} & \ding{52} & & \ding{52} & 38.1 & 75.3 & 63.9 & 94.6 \\
\ding{52} & & \ding{52} & \ding{52} & 38.2 & 75.1 & 63.6 & 95.3 \\
& \ding{52} & \ding{52}  & \ding{52} & 37.1 & 74.4 & 63.1 & 94.2 \\
\ding{52} & & & \ding{52}  & 36.5 & 73.6 & 62.9 & 94.4 \\
\ding{52} & \ding{52} & \ding{52} &  &  38.9 & 75.5 & 64.0 & 95.2 \\
\bottomrule
\end{tabular}
}


\end{table}

\paragraph{Consistency Loss Analysis.}
We examine the benefits of our fully learned domain affinity matrix in the consistency loss compared to a handcrafted baseline and the approach proposed by Yao \etal~\cite{yao2024improving}. More precisely, we implement and compare three variations: (\textbf{a}) fixed handcrafted weights defined by row-wise softmin of the angular distances between geographical coordinates; (\textbf{b}) a learned function that takes angular distances as input, which is similar to D\textsuperscript{3}G~\cite{yao2024improving}; and (\textbf{c}) a direct, unconstrained parameterization of the affinity matrix followed by a row-wise softmax normalization. We train our multi-head architecture with each weighting strategy and then perform domain expert selection, reporting the performance in \cref{tab:clexp}. Our unconstrained parameterization outperforms both the fixed angular weights and the adapted D\textsuperscript{3}G weighting scheme without requiring geographic metadata, indicating that inter-domain similarity may extend beyond purely geographic distance.

\paragraph{Impact of Backbone.}
In \cref{fig:backbone}, we present the performance gains achieved by our method over the baseline on all four datasets for three different backbones: MultiUTAE (our default), ResNet-10~\cite{gong2022resnet10}, and ResNet-18~\cite{he2016deep}. Because ResNet-10 and ResNet-18 are designed for single-image processing rather than spatio-temporal image time series (SITS), we apply them image-by-image. We observe that our method consistently improves performance across all datasets and all three backbones.

\paragraph{Efficiency.} 
Employing CoDEx with 55 domains (DynamicEarthNet) introduces approximately 213K additional parameters compared to baseline model, increasing training time by roughly 14\% (3.1\,min \vs 2.7\,min per epoch). At inference, the overhead remains minimal, around 6\% ($7.1 \times 10^{-3}$\,s \vs $6.7 \times 10^{-3}$\,s per $10^6$ pixels processed).
\begin{table}[t]
\caption{\textbf{Performance comparison of different affinity matrix formulations.} We compare three approaches: fixed, D\textsuperscript{3}G-style, and fully learned weights. Results are shown for both DynamicEarthNet and MUDS datasets. The \textbf{bold} values indicate the highest performance, while the \underline{underlined} values represent the second-highest.}
\centering
\small{
\begin{tabular}{lcccc}
\toprule
\multirow{2}{*}[-1mm]{Affinity} 
& \multicolumn{2}{c}{\textbf{DynEarthNet}} & \multicolumn{2}{c}{\textbf{MUDS}} \\  
\cmidrule(lr){2-3} \cmidrule(lr){4-5}
& {mIoU} & {OA} & {mIoU} & {OA} \\ 
\midrule
{\bf a)} Handcrafted & 38.8 & \underline{75.5} & 63.7 & 95.1 \\  
{\bf b)} ``D\textsuperscript{3}G~\cite{yao2024improving}-style'' & \underline{39.0} & 75.4 & \underline{64.0} & \underline{95.4} \\
{\bf c)} Learned (ours) & \textbf{39.1} & \textbf{75.7} & \textbf{64.2} & \textbf{95.8} \\    
\bottomrule
\end{tabular}
}

\label{tab:clexp}
\end{table}
\section{Conclusion}
We introduced {CoDEx}, a new framework for domain generalization in satellite imagery, addressing a challenging problem in Earth observation. Our key insight is to train multiple expert models, each specialized for a given domain of the training set, enforcing consistency among them and learning to select the most suitable expert for unseen domains. We validated our approach on four satellite image and time-series benchmarks across three tasks, outperforming ten state-of-the-art methods in both spatial domain generalization and adaptation.

\paragraph{Acknowledgement.}
    This work was supported by the European Research Council (ERC project DISCOVER, number 101076028) and by ANR project sharp ANR-23-PEIA-0008 in the context of the PEPR IA. This work was granted access to the HPC resources of IDRIS under the allocation 2024-AD011015600.


{\small
\bibliographystyle{templates/ICCV/ieee_fullname}
\bibliography{mybib}
}



\ARXIV{}{

}

\end{document}